\documentclass{article}

\usepackage{arxiv}

\usepackage[utf8]{inputenc} 
\usepackage[T1]{fontenc}    
\usepackage{hyperref}       
\usepackage{url}            
\usepackage{booktabs}       
\usepackage{amsfonts}       
\usepackage{nicefrac}       
\usepackage{microtype}      
\usepackage{lipsum}
\usepackage{graphicx}

\usepackage{amsmath} 
\usepackage{amssymb}
\usepackage{cleveref}
\usepackage{pifont}
\usepackage[dvipsnames,table,xcdraw]{xcolor}
\usepackage{subcaption}
\usepackage{multirow}
\usepackage{algorithm}
\usepackage{algorithmic}
\usepackage{booktabs}

\newcommand{\pub}[1]{{\color{gray}{\footnotesize{[{#1}]}}}}

\graphicspath{ {./images/} }

\title{Class Similarity-Based Multimodal Classification under Heterogeneous Category Sets}

\author{
  Yangrui Zhu\thanks{Equal contribution.} \\
  School of Computer Science\\ 
  Wuhan University\\
  Wuhan, China\\
  \texttt{yangruizhu@whu.edu.cn}
  \And
  Junhua Bao\footnotemark[1] \\
  School of Computer Science\\ 
  Wuhan University\\
  Wuhan, China\\
  \texttt{junhuabao@whu.edu.cn}
  \And
  Yipan Wei \\
  School of Computer Science\\ 
  Wuhan University\\
  Wuhan, China\\
  \texttt{yipanwei@whu.edu.cn}
  \And
  Yapeng Li\thanks{Corresponding author.}  \\
  School of Computer Science\\ 
  Wuhan University\\
  Wuhan, China\\
  \texttt{yapengli@whu.edu.cn}
  \And
  Bo Du\footnotemark[2] \\
  School of Computer Science\\ 
  Wuhan University\\
  Wuhan, China\\
  \texttt{bodu@whu.edu.cn}
}

\begin{document}
\maketitle
\begin{abstract}
Existing multimodal methods typically assume that different modalities share the same category set. However, in real-world applications, the category distributions in multimodal data exhibit inconsistencies, which can hinder the model's ability to effectively utilize cross-modal information for recognizing all categories.
In this work, we propose the practical setting termed Multi-Modal Heterogeneous Category-set Learning (MMHCL), where models are trained in heterogeneous category sets of multi-modal data and aim to recognize complete classes set of all modalities during test. To effectively address this task, we propose a Class Similarity-based Cross-modal Fusion model (CSCF). 
Specifically, CSCF aligns modality-specific features to a shared semantic space to enable knowledge transfer between seen and unseen classes. It then selects the most discriminative modality for decision fusion through uncertainty estimation. Finally, it integrates cross-modal information based on class similarity, where the auxiliary modality refines the prediction of the dominant one.
Experimental results show that our method significantly outperforms existing state-of-the-art (SOTA) approaches on multiple benchmark datasets, effectively addressing the MMHCL task.
\end{abstract}

\section{Introduction}

In real-world applications, due to limitations in equipment, time costs, and budget, datasets are typically collected for specific tasks. For example, a project focused on video animal classification may only collect video data, while a project developing an audio action recognition system may primarily collect audio data. However, from a multimodal learning perspective, both video and audio data can inherently provide valuable information for these recognition tasks. This leads to a critical research question: \textbf{Can a unified multimodal recognition model be constructed by integrating modality data from heterogeneous category sets?} Such a model needs not only to handle all relevant modality data but also to recognize the complete category set. 

\begin{figure}[!t]
    \centering
    \begin{subfigure}{\columnwidth}
        \centering
        \includegraphics[width=0.7\linewidth]{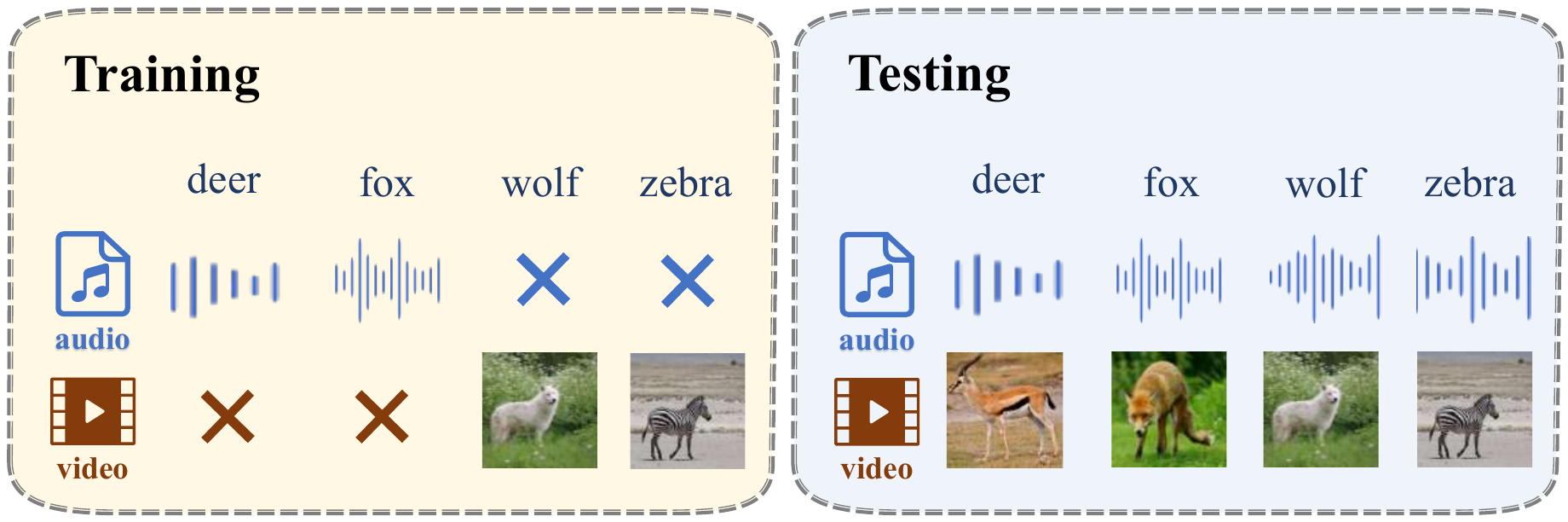}
        \caption{Our Setting}
        \label{fig:setting}
    \end{subfigure}

    \vspace{8pt}

    \begin{subfigure}{\columnwidth}
        \centering
        \begin{tabular}{c|cccc}

        \hline
        Dataset     & DGZ~\cite{chen2023deconstructed}   & TMC~\cite{han2020trusted} & UVaT~\cite{10666988}  & Ours                     \\ \hline
        ActivityNet & 31.11 & 32.72 & 35.48 & \textbf{45.71
        } \\ 
        \hline
        \end{tabular}
        \caption{Results}
        \label{tab:results}
    \end{subfigure}

    \caption{Setting illustration and results: (a) Setup for MMHCL. During the training phase, each modality has an inconsistent category set, with \ding{53} indicating categories not covered by each modality. In the testing phase, the model needs to recognize the complete category space across all modalities. (b) Performance comparison on the ActivityNet dataset 
    (Real-world Scenarios)
    under MMHCL scenario. Our method significantly outperforms existing SOTA approaches.}
    \label{fig:combined setting and results}
\end{figure}

To address this, we propose a new paradigm called Multi-Modal
Heterogeneous Category-set Learning (MMHCL, as shown in~\Cref{fig:combined setting and results}a).

\begin{figure*}[!t]
    \centering
    \includegraphics[width=\linewidth]{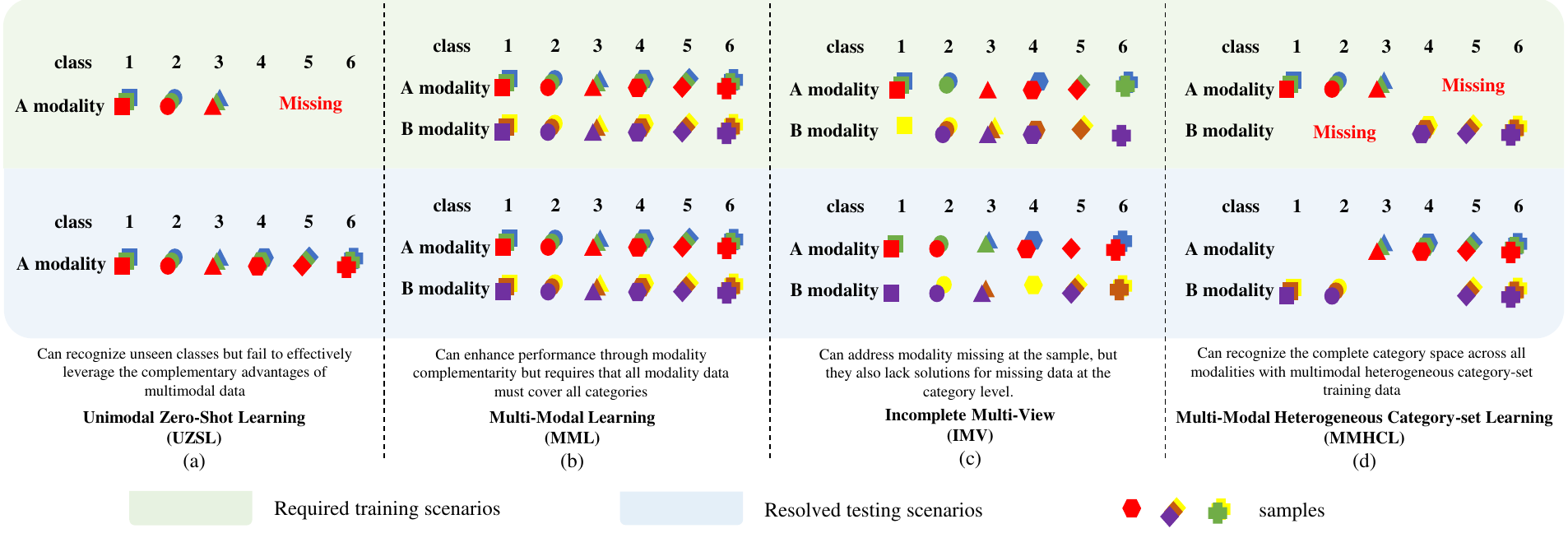}
     \caption{Comparison of Settings. Unimodal Zero-shot learning (UZSL) focuses on seen classes and unseen classes within a single modality, making it unsuitable for cross-modal scenarios. Multimodal learning (MML) requires all modality data to cover all categories during training, and testing is conducted on complete samples with seen classes. Incomplete view (IMV) classification allows for modalities missing at the sample during both training and testing but lack solutions for missing data at the category level. Under our proposed setting (MMHCL), the training data exhibits inconsistent class coverage across modalities, and the model is required to recognize the complete category space across all modalities.}
    \label{fig:setting_comparison}
\end{figure*}

Current related research has three main limitations: First, traditional multimodal classification methods~\cite{han2020trusted,xu2024trusted} can enhance performance through modality complementarity, but their basic assumption requires that all modality data must cover all categories (as shown in~\Cref{fig:setting_comparison}b), making them incapable of handling data missing at the category level. Second, incomplete multi-view classification methods~\cite{liu2023incomplete,10666988, acmmm_imv1, acmmm_imv2} address modality missing at the sample level (as shown in~\Cref{fig:setting_comparison}c), but they also lack solutions for missing data at the category level. Third, zero-shot learning methods enable recognition of unseen categories through generation mechanisms or feature alignment, but they fail to effectively leverage the complementary advantages of multimodal data~\cite{chen2023deconstructed,li2024improving, acmmm_zsl1} or requires that all modality data must be complete~\cite{mazumder2021avgzslnet,mercea2022audio,hong2023hyperbolic,kurzendorfer2024audio}. 

Therefore, we aim to develop a method that can effectively enhance robustness in addressing the MMHCL task.

To achieve this goal, we are confronted with three main challenges: 1) How to enable unimodal data, which only includes part of the categories, to recognize the complete category set including unseen classes; 2) In a multimodal prediction scenario, how to accurately identify the modality domain of the data and allow the corresponding modality to dominate the fusion decision; 3) How to effectively utilize complementary information from other modalities to assist the dominant modality's prediction. To address these challenges, we propose a Class Similarity-based Cross-modal Fusion model (CSCF), which is trained on modality-specific data with heterogeneous category sets and is capable of recognizing the complete category space across all modalities. Specifically, we enhance category semantic representation through large language models (LLM), constructing a unified semantic space as a bridge between seen and unseen categories. We map data from each modality to this semantic space and leverage the semantic correlations between categories to facilitate knowledge transfer, enabling generalization to unseen categories (\textbf{challenge 1}). 

Furthermore, we design uncertainty estimation to assess prediction inconsistencies among sub-modules and differences across modalities in a dynamic and robust manner, thus determining the dominant modality and ensuring that the fusion decision is driven by the most discriminative modality (\textbf{challenge 2}). Finally, guided by class semantic similarity, we adaptively fuse complementary information from other modalities to enhance overall prediction capability (\textbf{challenge 3}).

To summarize, the contributions of this paper are as follows:
\begin{itemize}
\item To the best of our knowledge, this is the first study on MMHCL, which utilizes multimodal data with different category sets to develop a model capable of recognizing all categories across all modalities.
\item We propose a Class Similarity-based Cross-modal Fusion model (CSCF), which requires only one training session and can simultaneously integrate cross-modal information and recognize complete category space across all modalities.
\item We find that leveraging semantic discriminability, modality prediction consistency, and class semantic similarity is crucial to solving this problem.
\end{itemize}
We conduct extensive experiments on diverse real-world datasets. The results show that in the practical MMHCL setting, our method significantly outperforms the most competitive counterpart by $10.23\%$, $10.77\%$, and $3.30\%$ on the ActivityNet, UCF, and VGGSound datasets, respectively.

\section{Related Work}
\subsection{Multimodal Learning}
To fully utilize the information from different modalities, multimodal learning~\cite{ramachandram2017deep,baltruvsaitis2018multimodal,wang2021survey,sleeman2022multimodal, acmmm_mml1, acmmm_mml2, acmmm_mml3, acmmm_mml4} has become a widely researched field. Based on different fusion strategies, multimodal learning methods can generally be divided into three categories: early fusion~\cite{poria2015deep,garillos2021multimodal}, intermediate fusion~\cite{arevalo2017gated,kiela2019supervised,tsai2019multimodal,hong2020more,huang2020multimodal,wang2020deep,hu2021unit,huang2021makes,lee2021variational,li2021multi}, and decision fusion~\cite{natarajan2012multimodal,subedar2019uncertainty,han2020trusted,han2022multimodal,xu2024trusted,huang2024trusted}. Early fusion based methods directly integrate multiple modalities at the data level, typically concatenating multimodal data.
Intermediate fusion combines data from different modalities at intermediate layers of a neural network~\cite{arevalo2017gated,kiela2019supervised,tsai2019multimodal,hong2020more,huang2020multimodal,wang2020deep,hu2021unit}, effectively capturing the complementary information of each modality to enhance the model's performance. Decision fusion typically involves the integration of multimodal data based on uncertainty evaluation~\cite{gal2016dropout,sensoy2018evidential,lyzhov2020greedy}, voting schemes~\cite{morvant2014majority}, and so on~\cite{shutova2016black,1230212,glodek2011multiple,ramirez2011modeling}.
However, these multimodal methods assume that all modalities must cover all categories and use fully connected layers as classifiers. They only recognize the seen categories for each modality and cannot generalize to the unseen categories for each modality. In contrast, our approach uses semantics as a bridge to map data from each modality into a semantic space, enabling the recognition of unseen categories in each modality.

\subsection{Incomplete Multi-View Classification}
Incomplete multi-view classification has been widely studied to address the common problem of missing modality in some samples in real-world scenarios. Existing incomplete multi-view (IMV) learning methods can be roughly divided into two categories: model-independent methods~\cite{jaques2015multi,Tran_2017_CVPR,8257992} and model-based methods~\cite{JIANG2021106,NEURIPS2019_11b9842e,liu2023incomplete,10666988,9447974}. Model-independent methods generally involve a two-stage process to tackle the incomplete multi-view problem. In the first stage, a model-independent preprocessing strategy is designed (e.g., imputation~\cite{Tran_2017_CVPR,8257992} or simply discarding samples with missing views) to construct a complete multi-view dataset. In the second stage, standard multi-view learning methods are applied to process the reconstructed complete data. On the other hand, model-based methods address the incomplete views problem by designing specific multi-view learning models based on traditional machine learning or deep learning. For example, LMVCAT~\cite{liu2023incomplete} uses label-guided masked views and a category-aware transformer to avoid the negative impact of missing views on information interaction, while UVaT~\cite{10666988} employs a view-aware transformer to mask the missing views and adaptively explores the relationship between views and target tasks to handle missing views.
However, these methods only consider modality missing at the sample level (some samples may lack certain modalities) but overlook the modality-category missing problem that arises in some scenarios (where certain modalities completely lack specific categories). This limitation causes a significant decline in model generalization when the test data contains categories unseen by a specific modality during training. In contrast, our method effectively solves the more challenging modality-category missing problem.

\subsection{Generalized Zero-Shot Learning}

Generalized Zero-Shot Learning (GZSL) can be broadly categorized into unimodal and multimodal approaches.
In unimodal settings, the predominant methods fall into two paradigms: embedding-based\\~\cite{8003482,8476580,Li_2017_CVPR,zhang2019co,skorokhodov2020class,chen2022msdn,chen2023zero,li2024improving} and generation-based~\cite{8957359,Paul_2019_CVPR,Wu_2020_CVPR,xian2018feature,schonfeld2019generalized,narayan2020latent,chen2021free,chen2022zero,chen2023deconstructed,chen2023evolving}.
Embedding-based methods aim to learn a shared embedding space that aligns visual and semantic features. By measuring similarity between the prototype of a seen class and the projected representation of a predicted class in this space, the model can recognize unseen categories.
Generation-based methods, on the other hand, employ generative models to synthesize visual features or even images for unseen classes, conditioned on their semantic descriptions. This effectively transforms the GZSL problem into a standard supervised classification task.
However, these unimodal methods fail to exploit the complementary information present in multimodal data. To address these challenges, researchers have proposed multimodal GZSL methods~\cite{mazumder2021avgzslnet,mercea2022audio,hong2023hyperbolic,kurzendorfer2024audio}. For example:
AVCA~\cite{mercea2022audio} employs a cross-attention mechanism to enable effective multimodal information fusion.
CLIPCLAP~\cite{kurzendorfer2024audio} leverages high-quality features extracted from large-scale pretrained models such as CLIP and CLAP to enhance performance.
Despite their advances, most existing methods rely on an idealized assumption:
That multimodal data must be complete, which is often invalid in real-world applications where different modalities may cover distinct subsets of the category set.
In contrast, our proposed method focuses on more realistic scenarios with modality incompleteness and category-set inconsistency, thereby improving the model’s practicality.

\section{Method}

\textbf{Problem Formulation.}
In the MMHCL setup, we consider $m$ modalities, denoted as $\mathcal{M} = \{M_1, M_2, \dots, M_m\}$. For each modality $M_i$, we define its seen category set as $Y_s^i$ and unseen category set as $Y_u^i$. The corresponding data and labels are denoted as $X^i$ and $y^i$, respectively.
In practice, each modality covers only a subset of the total category space, and the seen category sets across modalities may partially overlap or be completely disjoint. Formally, we have:
$\bigcup_{i=1}^{m} Y_s^i = \mathcal{Y}, \quad \text{and} \quad Y_s^i \cap Y_s^j \neq \varnothing \quad \text{or} \quad Y_s^i \cap Y_s^j = \varnothing, \quad \forall i \ne j,$
where $\mathcal{Y}$ denotes the union of all categories involved across modalities.
Each training sample is represented as $\{X_k^i, y_k\}$, where $X_k^i \in X^i$ and $y_k \in \mathcal{Y}$, indicating that the $k$-th sample only exists in modality $M_i$, and data from other modalities is considered missing.
Our objective is to train an end-to-end multimodal classifier that can recognize the complete category space across all modalities, given all modality-specific training data $\{X^1, \dots, X^m\}$ and their corresponding labels $\{y^1, \dots, y^m\}$.
This work focuses on a fundamental yet challenging case with two modalities, where each modality corresponds to a disjoint set of categories.

\begin{figure*}[htb]
    \centering
    \includegraphics[width=\linewidth]{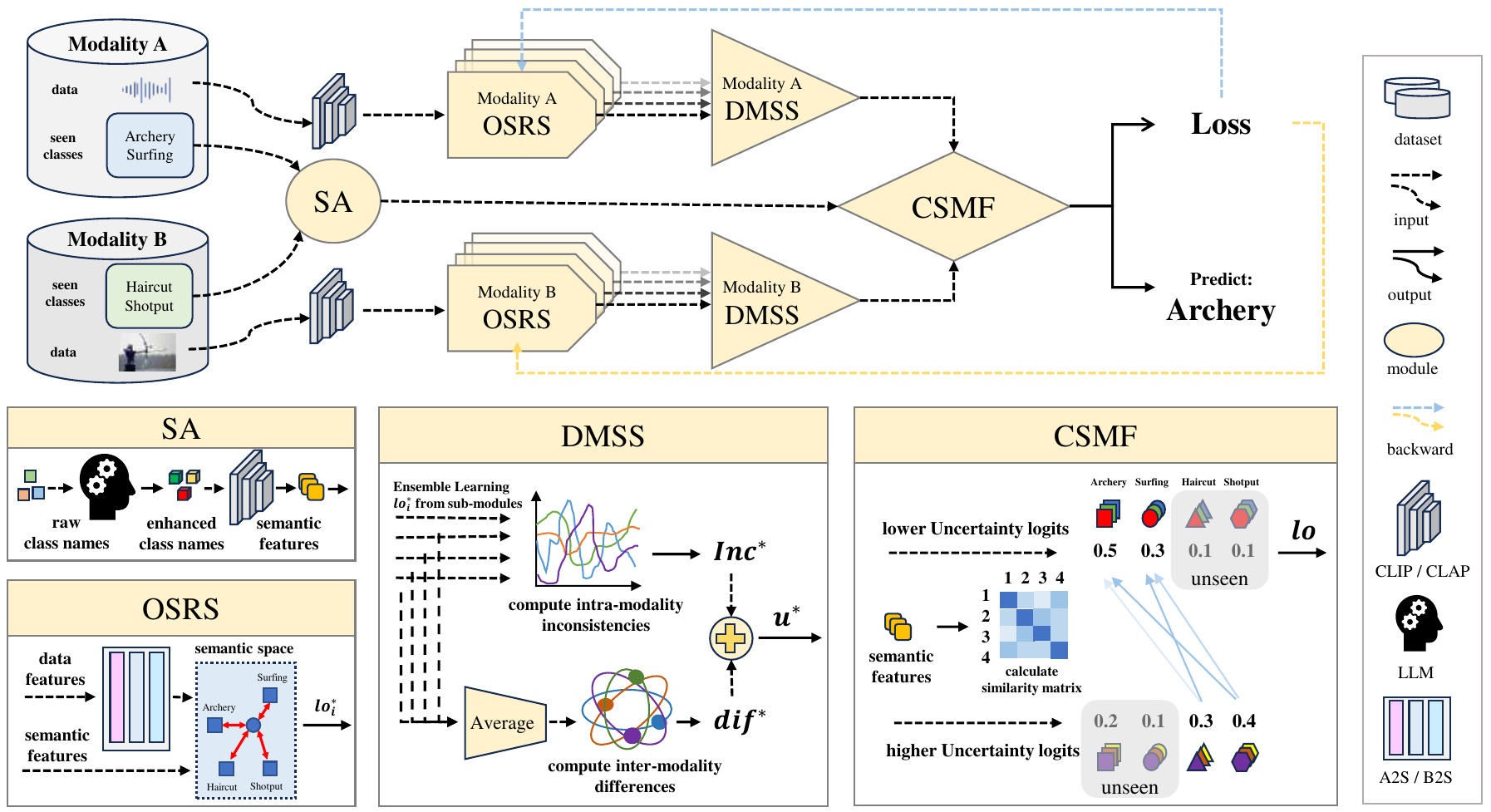}
    \caption{Overall framework of the proposed CSCF model. In the data processing phase, semantic enhancement is performed based on LLMs, and features are extracted using CLIP/CLAP. During training, our approach consists of three main modules: OSRS, DMSS, and CSMF. In the OSRS module, data features are mapped into the semantic space and classified based on distance. In the DMSS module, we consider both intra-modality inconsistencies and inter-modality differences of predictions, selecting the dominant modality to lead the prediction. In the CSMF module, we adaptively fuse multimodal predictions using class semantic similarity, where auxiliary modalities refine the dominant modality’s prediction. During inference, the model leverages all three modules to recognize the complete category space across all modalities.}
    \label{fig:Method}
\end{figure*}

\noindent\textbf{Overall Framework.}
The proposed CSCF model is illustrated in~\Cref{fig:Method}.
From a unimodal perspective, the model is trained on seen classes and must generalize to unseen ones. To bridge this gap, we enhance class semantics using large language models (LLMs)~\cite{achiam2023gpt} and map features into a shared semantic space (see the OSRS module in~\Cref{fig:Method}).
From a multimodal perspective, each modality covers only part of the category space, leading to varying prediction quality. 
So we design uncertainty estimation to identify the dominant modality for decision-making (see the DMSS module in~\Cref{fig:Method}).
Finally, we perform adaptive fusion guided by class semantic similarity, where auxiliary modality predictions refine the dominant modality’s output for comprehensive and accurate recognition (see the CSMF module in~\Cref{fig:Method}).

\subsection{Open Set Recognition via Semantics}
\label{sec:OSRS}

\textbf{Motivation of OSRS.}
In real-world scenarios, each modality only covers a subset of categories, requiring the model to recognize unseen classes from a unimodal perspective. We introduce class semantics as a bridge between seen and unseen classes. To enable knowledge transfer, we enhance semantic features using large language models (LLMs) and adopt the concept of open set recognition~\cite{geng2020recent}. Modality-specific semantic mappers project features into a unified semantic space, allowing the model to leverage semantics for unseen class recognition.

\noindent\textbf{Construction of OSRS.}
We use the semantic space as the shared alignment space. The feature $\mathbf{x}^A$ from modality A is mapped to the semantic space via a mapper $A2S$, while the feature $\mathbf{x}^B$ from modality B is mapped through $B2S$. The mapping process is defined as follows: 
\begin{equation}
\begin{split}
    \label{eq:mapper}
    \mathbf{x}^A \in \mathbb{R}^{N^A \times d^{A}} \xrightarrow{A2S} \hat{\mathbf{s}}^A \in \mathbb{R}^{N^A \times d^{s}}, \\
    \mathbf{x}^B \in \mathbb{R}^{N^B \times d^{B}} \xrightarrow{B2S} \hat{\mathbf{s}}^B \in \mathbb{R}^{N^B \times d^{s}},
\end{split}
\end{equation}
where $d^A$ and $d^B$ are the input feature dimensions for the two modalities, and $d^s$ is the output dimension aligned with the semantic feature space. $\hat{\mathbf{s}}^A$ and $\hat{\mathbf{s}}^B$ are the semantic features derived from the features of each modality. $N^A$ and $N^B$ denote the number of samples from seen classes in modalities A and B, respectively.

Next, we compute the scaled cosine similarity between the modality-generated semantic features $\hat{\mathbf{s}}^*$ and the enhanced semantic features $\mathbf{c}_i$, where $\mathbf{c}_i$ represents the enhanced semantic feature of the $i$-th class name (see appendix A for implementation details). The resulting similarity-based predictions $\mathbf{lo}^*$ are computed as:
\begin{equation}
\begin{split}
    \label{eq:compute cosine similarity}
    &lo_i^* = \frac{\gamma^2\hat{\mathbf{s}}^*\mathbf{c}_i}{\|\hat{\mathbf{s}}^*\|\left\|\mathbf{c}_i\right\|}, \\
    &\mathbf{lo}^* = [lo_1^*, lo_2^*, \dots, lo_N^*],
\end{split}
\end{equation}
where $*$ denotes either modality A or B.

\subsection{Dominant Modality Selection Strategy}
\label{sec:DMSS}

\textbf{Motivation of DMSS.}
In multimodal settings, each modality covers only part of the category space, making it difficult to achieve high-quality recognition individually, which affects the final fusion. Thus, the model must assess the reliability of each modality. Inspired by ensemble learning~\cite{sagi2018ensemble}, we design uncertainty estimation to evaluate prediction quality. Empirically, OSRS modules within a modality show low inconsistency on seen classes and high inconsistency on unseen ones (see Appendix B). Based on this, we propose an entropy-based mechanism that measures intra-modality inconsistencies and inter-modality differences of predictions, enabling the model to dynamically select the dominant modality for robust fusion decisions.

\noindent\textbf{Consturction of DMSS.}
We apply ensemble learning by aggregating predictions from multiple OSRS modules and converting them into probability distributions: 
\begin{equation}
\begin{split}
\label{eq:ensemble OSRS}
&\mathbf{lo}_i^*=OSRS_i(\mathbf{x}^*, \mathbf{c}), \\
&\mathbf{p}_i^* = softmax(\mathbf{lo}_i^*),
\end{split}
\end{equation}
where $*$ denotes modality A or B, $i$ refers to the $i$-th OSRS module and $\mathbf{c}$ represents for the semantic features of all classes. When predictions from multiple OSRS modules within a modality are lowly inconsistent, it indicates that the modality can provide stable and reliable decisions for the current sample.
To measure intra-modality disagreement of predictions, we use entropy-based inconsistency: 
\begin{equation}
\begin{split}
\label{eq:entropy-based inconsistency}
e^* = \sqrt{\frac{1}{K} \sum_{i=1}^K \left( H(\mathbf{p}_i^*) - \bar{H} \right)^2 },
\end{split}
\end{equation}
where $H(\mathbf{p}_i^*) = - \sum_{n=1}^N p_{i,n}^* \log p_{i,n}^*$ is the entropy of the $i$-th module’s prediction, and $\bar{H} = \frac{1}{K} \sum_{i=1}^K H(\mathbf{p}_i^*)$ is the mean entropy. The intra-modality inconsistency $Inc^*$ for modality $*$ is then computed as: 
\begin{equation}
\begin{split}
\label{eq:intra-modality inconsistency}
Inc^{*} = \frac{e^*}{\sum_{* \in \left\{A,\ B\right\}}e^* },  \quad * \in \left\{A,\ B\right\}.
\end{split}
\end{equation}

In addition, if the prediction from one modality significantly deviates from the others, it may indicate that the modality has captured critical discriminative information. To quantify this, we compute inter-modality prediction differences using entropy-based uncertainty estimation:
\begin{equation} 
\begin{split} 
\label{eq:inter-modality differences} 
&\mathbf{p}^{*} = softmax(\frac{1}{K}\sum_{i=1}^{K}\mathbf{lo}_{i}^{*}), \\
&dif^{*} = \frac{H(\mathbf{p}^*)}{\sum_{* \in \left\{A,\ B\right\}}H(\mathbf{p}^*) },  \quad * \in \left\{A,\ B\right\}.
\end{split} 
\end{equation}

Finally, we combine intra-modality inconsistency and inter-modality prediction difference to compute the overall prediction uncertainty $u^*$ for each modality: \begin{equation} 
\begin{split} 
\label{eq:overall prediction uncertainty} 
u^*= Inc^*+dif^*,
\end{split} 
\end{equation}
where $*$ denotes modality A or B. The dominant modality usually corresponds to the one with the lower uncertainty (refer to~\Cref{fig:uncertainty analysis}). By comparing $u^A$ and $u^B$, the model can identify the dominant and auxiliary modalities, thereby enhancing the robustness of the final fusion decision.

\subsection{Class Similarity-guided Multimodal Fusion}
\label{sec:CSMF}

\textbf{Motivation of CSMF.}
Different modalities vary in their ability to recognize certain classes. For example, the audio modality may better capture the neighing of a horse, while the video modality excels at identifying a donkey’s appearance. If the audio modality predicts “horse” and is selected as dominant by the DMSS module, the video prediction is still valuable due to the semantic similarity between “horse” and “donkey.” Thus, the auxiliary (video) modality’s prediction should be reasonably integrated to assist the dominant (audio) modality in making a more accurate decision.
Motivated by this, we propose an adaptive fusion strategy guided by class semantic similarity. Predictions from the auxiliary modality are weighted based on their semantic relevance to the dominant modality’s prediction and then fused to improve overall classification. This mechanism not only leverages the strength of the dominant modality but also captures complementary information from auxiliary modalities.

\noindent\textbf{Construction of CSMF.}
We concatenate the enhanced class semantic features $\mathbf{c}_i$ to form the class semantic feature matrix $\mathbf{C}$, and compute cosine similarity between each pair of class semantics to obtain a similarity matrix $\mathbf{S} \in \mathbb{R}^{N \times N}$, where $N$ is the number of classes: 
\begin{equation}
\begin{split}
\label{eq:compute class similarity}
&\mathbf{C} = [\mathbf{c}_1, \mathbf{c}_2, \dots, \mathbf{c}_N]^\top \in \mathbb{R}^{N \times d^s}, \\
&S_{i,j} = \cos(\mathbf{c}_i, \mathbf{c}_j) = \frac{\mathbf{c}_i^\top \mathbf{c}_j}{\|\mathbf{c}_i\|_2 \cdot \|\mathbf{c}_j\|_2}, \quad \forall i,j \in \{1, \dots, N\}.
\end{split}
\end{equation}

For modality $*$, we retain only the $top$-$k$ most semantically similar classes for each seen class, forming a pruned similarity matrix $\mathbf{S}^*$. The auxiliary modality’s predictions are then re-weighted according to this similarity matrix and added to the dominant modality’s predictions to yield the final prediction $\mathbf{lo}$: 
\begin{equation}
\begin{split}
\label{eq:final prediction}
&\tilde{\mathbf{lo}}^A = \mathbf{S}^A \cdot \mathbf{lo}^A, \quad \tilde{\mathbf{lo}}^B = \mathbf{S}^B \cdot \mathbf{lo}^B, \\
&\mathbf{lo} =
\begin{cases}
\mathbf{lo}^A + \tilde{\mathbf{lo}}^B, & \text{if } \mathrm{u}^A < \mathrm{u}^B, \\
\mathbf{lo}^B + \tilde{\mathbf{lo}}^A, & \text{otherwise},
\end{cases}
\end{split}
\end{equation}
where $\tilde{\mathbf{lo}}^A$ and $\tilde{\mathbf{lo}}^B$ are the similarity-enhanced components, and the fusion decision is dynamically made based on the prediction uncertainty $u^*$ estimated by the DMSS module. Finally, the fused logits are converted into a probability distribution, enabling high-quality recognition across all categories.

\begin{algorithm}[!t]
   \caption{Training Process of CSCF}
   \label{alg:training process}
\begin{algorithmic}
   \STATE {\bfseries Input:} Training data $X^A, Y_s^A\}$ and $\{X^B, Y_s^B\}$ from modalities A and B, semantic class features $\{\mathbf{c}_i\}_{i=1}^N$
   \STATE {\bfseries Output:} Trained parameters of $A2S$, $B2S$.

   \STATE \textcolor[RGB]{61,102,150}{\small{// Feature Extraction and Semantic Mapping}}
   \STATE $\mathbf{x}^A = \text{Backbone}_A(X^A), \quad \mathbf{x}^B = \text{Backbone}_B(X^B)$
   \STATE $\mathbf{c}=CLIP(LLM(Y))$

   \FOR{$e=1$ {\bfseries to} $E$}
    \STATE $\hat{\mathbf{s}}^A = A2S(\mathbf{x}^A), \quad \hat{\mathbf{s}}^B = B2S(\mathbf{x}^B)$
        \STATE \textcolor[RGB]{61,102,150}{\small{{// Ensemble OSRS: compute logits and probs}}}
        \FOR{$i=1$ {\bfseries to} $K$}
            \STATE $\mathbf{lo}_i^A, \mathbf{p}_i^A = OSRS_i(\hat{\mathbf{s}}^A, \mathbf{c}), \mathbf{lo}_i^B, \mathbf{p}_i^B = OSRS_i(\hat{\mathbf{s}}^B, \mathbf{c})$
        \ENDFOR
        \STATE $\mathbf{p}^A, \mathbf{p}^B \gets \{(\mathbf{lo}_1^A, \dots, \mathbf{lo}_K^A), (\mathbf{lo}_1^B, \dots, \mathbf{lo}_K^B)\}$ in Eq.(\ref{eq:ensemble OSRS})

        \STATE \textcolor[RGB]{61,102,150}{\small{// DMSS: Compute modality uncertainty}}
        \STATE $Inc^A, Inc^B \gets \{\mathbf{p}_1^A, \dots, \mathbf{p}_K^A; \mathbf{p}_1^B, \dots, \mathbf{p}_K^B\}$ in Eq.(\ref{eq:entropy-based inconsistency})
        \STATE $dif^A, dif^B \gets \{\mathbf{p}^A, \mathbf{p}^B\}$ in Eq.(\ref{eq:inter-modality differences})
        \STATE $u^A = Inc^A + dif^A, \quad u^B = Inc^B + dif^B$

        \STATE \textcolor[RGB]{61,102,150}{\small{// CSMF: Similarity-guided prediction fusion}}
        \STATE $\mathbf{S}_{i,j} = \frac{\mathbf{c}_i^\top \mathbf{c}_j}{\|\mathbf{c}_i\| \cdot \|\mathbf{c}_j\|}$
        \STATE $\mathbf{lo} \gets \{lo^A, lo^B, \mathbf{S}^A, \mathbf{S}^B\, u^A, u^B\}$ in Eq.(\ref{eq:final prediction})

        \STATE \textcolor[RGB]{61,102,150}{\small{// Compute total loss and update parameters}}
        \STATE $\mathcal{L} = \sum_{* \in \{A, B\}}(\frac{1}{K}\sum_{i=1}^{K}CE(\mathbf{p}_{i}^{*},y) + CE(\mathbf{p}^*,y))$
        \STATE Update parameters $\{\theta_{\text{A2S}_i, \text{B2S}}\}_{i=1}^{K}$ using Adam
   \ENDFOR
\end{algorithmic}
\end{algorithm}

\subsection{Training and Inference}

During the training phase, the overall training objective $\mathcal{L}$ is defined as: 
\begin{equation} 
\label{eq:unimodal loss} 
\mathcal{L} = \sum_{* \in \{A, B\}}(\frac{1}{K}\sum_{i=1}^{K}CE(\mathbf{p}_{i}^{*},y) + CE(\mathbf{p}^*,y)), 
\end{equation}
where $*$ denotes modality A or B, $K$ is the number of OSRS modules, $CE$ represents the cross-entropy loss function, $\mathbf{p}_{i}^{*}$ is the prediction probability from the $i$-th OSRS module for modality $*$, $\mathbf{p}^{*}$ is the average probability computed as in Equation (\ref{eq:inter-modality differences}), and $y$ is the ground truth label.
The complete training process is illustrated in~\Cref{alg:training process}. At the end of each training epoch, the parameters of the A2S and B2S mappers are updated.

During inference, for each input multimodal sample, features are first extracted and then processed similarly to the training phase. The OSRS modules map each modality’s features into the semantic space via the corresponding mapper and computes similarity scores with all class semantics in $Y$ to generate per-modality predictions.
The DMSS module aggregates the predictions from multiple OSRS modules to evaluate intra-modality inconsistencies and inter-modality differences, thereby determining the dominant modality for fusion.
Finally, the CSMF module performs adaptive multimodal fusion based on class semantic similarity, yielding the final prediction result.

\begin{table*}[t]
\caption{Comparison with SOTA methods in generalization ability, multimodal fusion ability and real-world scenarios. $A_u$ and $B_u$ represent the test results for unseen classes in single modality; $A_s+B_u$, $A_u+B_S$ and $A_\text{all}+B_\text{all}$ represent the test results for different category sets in complete modalities; $acc_\text{mix}$ indicates the test results for mixed scenarios ($A_s$, $B_s$, $A_u$, $B_u$, $A_s+B_u$, $A_u+B_s$). For detailed descriptions of evaluation methods, please refer Appendix C.}
\label{tab:comparison method}
\resizebox{\textwidth}{!}{
\begin{tabular}{c|cc|cc|ccc|c}
\hline
\hline
 
 &
  \multicolumn{2}{c|}{} &
  \multicolumn{2}{c|}{Generalization Ability} &
  \multicolumn{3}{c|}{Multimodal Fusion Ability} &
  Real-world Scenarios \\
 
\multirow{-2}{*}{Dataset} &
  \multicolumn{2}{c|}{\multirow{-2}{*}{Methods}} &
\( A_u \) &
\( B_u \) &
\( A_s + B_u \) &
\( A_u + B_s \) &
\( A_{\text{all}} + B_{\text{all}} \) &
\( \text{acc}_{\text{mix}} \) \\ \hline
 
 & \multicolumn{1}{c|}{}                      & ZLA\pub{IJCAI22}~\cite{chen2022zero}        & \underline{2.04} & 4.14 & 8.89  & 64.45 & 38.35 & 29.74 \\
 
 & \multicolumn{1}{c|}{}                      & DGZ\pub{AAAI23}~\cite{chen2023deconstructed}       & 0.77 & 3.49 & 11.45 & 65.09 & 39.89 & 31.11 \\
 
 & \multicolumn{1}{c|}{\multirow{-3}{*}{ZSL}} & DSECN\pub{CVPR24}~\cite{li2024improving}      & 1.66 & \underline{8.10} & 23.41 & 54.24 & 39.76 & 30.62 \\ \cline{2-3}
 
 & \multicolumn{1}{c|}{}                      & TMC\pub{TPAMI22}~\cite{han2020trusted}        & 0.00 & 0.00 & 17.25 & \underline{69.11} & \underline{44.75} & 32.72 \\
 
 & \multicolumn{1}{c|}{\multirow{-2}{*}{MML}}                      & MMdynamics\pub{CVPR22}~\cite{han2022multimodal} & 0.00 & 0.00 & 17.46 & 32.42 & 24.14 & 23.52 \\ \cline{2-3}
 
 & \multicolumn{1}{c|}{}                      & LMVCAT\pub{AAAI23}~\cite{liu2023incomplete}     & 0.00 & 0.00 & 3.31  & 48.47 & 27.25 & 22.18 \\
 
 & \multicolumn{1}{c|}{\multirow{-2}{*}{IMV}} & UVaT\pub{TIP24}~\cite{10666988}       & 0.00 & 0.00 & \underline{41.30} & 33.82 & 37.34 & \underline{35.48} \\ \cline{2-3}
\multirow{-9}{*}{ActivityNet} &
  \multicolumn{2}{c|}{CSCF (Ours)} &
  \textbf{3.96} &
  \textbf{11.20} &
  \textbf{48.61} &
  \textbf{84.33} &
  \textbf{67.55} &
  \textbf{45.71} \\ \hline
 
 & \multicolumn{1}{c|}{}                      & ZLA\pub{IJCAI22}~\cite{chen2022zero}        & \underline{9.73} & 6.89 & 56.39 & \underline{74.62} & 65.48 & \underline{51.51} \\
 
 & \multicolumn{1}{c|}{}                      & DGZ\pub{AAAI23}~\cite{chen2023deconstructed}        & 4.81 & \underline{8.68} & \underline{70.36} & 54.76 & 48.02 & 50.74 \\
 
 & \multicolumn{1}{c|}{\multirow{-3}{*}{ZSL}} & DSECN\pub{CVPR24}~\cite{li2024improving}      & 4.31 & 6.69 & 68.46 & 40.12 & 54.33 & 44.91 \\ \cline{2-3}
 
 & \multicolumn{1}{c|}{}                      & TMC\pub{TPAMI22}~\cite{han2020trusted}        & 0.00 & 0.00 & 67.56 & 71.61 & \underline{69.58} & 51.13 \\
 
 & \multicolumn{1}{c|}{\multirow{-2}{*}{MML}}                      & Mmdynamics\pub{CVPR22}~\cite{han2022multimodal} & 0.00 & 0.00 & 61.88 & 45.34 & 52.78 & 47.89 \\ \cline{2-3}
 
 & \multicolumn{1}{c|}{}                      & LMVCAT\pub{AAAI23}~\cite{liu2023incomplete}     & 0.00 & 0.00 & 57.29 & 63.19 & 60.23 & 48.67 \\
 
 & \multicolumn{1}{c|}{\multirow{-2}{*}{IMV}} & UVaT\pub{TIP24}~\cite{10666988}       & 0.00 & 0.00 & 70.26 & 38.19 & 52.07 & 42.24 \\ \cline{2-3}
\multirow{-9}{*}{UCF} &
  \multicolumn{2}{c|}{CSCF (Ours)} &
  \textbf{9.83} &
  \textbf{14.97} &
  \textbf{86.23} &
  \textbf{93.08} &
  \textbf{89.64} &
  \textbf{62.28} \\ \hline
 
 & \multicolumn{1}{c|}{}                      & ZLA\pub{IJCAI22}~\cite{chen2022zero}        & 3.54 & \underline{6.04} & 48.10 & 41.92 & 44.59 & 34.99 \\
 
 & \multicolumn{1}{c|}{}                      & DGZ\pub{AAAI23}~\cite{chen2023deconstructed}        & \underline{4.93} & 3.24 & 60.55 & 43.37 & 50.80 & 39.86 \\
 
 & \multicolumn{1}{c|}{\multirow{-3}{*}{ZSL}} & DSECN\pub{CVPR24}~\cite{li2024improving}      & 4.43 & 2.94 & \underline{68.52} & 30.42 & 46.91 & 38.31 \\ \cline{2-3}
 
 & \multicolumn{1}{c|}{}                      & TMC\pub{TPAMI22}~\cite{han2020trusted}        & 0.00 & 0.00 & 64.44 & \underline{49.80} & \underline{56.14} & \underline{41.80} \\
 
 & \multicolumn{1}{c|}{\multirow{-2}{*}{MML}}                      & Mmdynamics\pub{CVPR22}~\cite{han2022multimodal} & 0.00 & 0.00 & 48.92 & 40.77 & 46.42 & 37.15 \\ \cline{2-3}
 
 & \multicolumn{1}{c|}{}                      & LMVCAT\pub{AAAI23}~\cite{liu2023incomplete}     & 0.00 & 0.00 & 46.14 & 49.11 & 47.83 & 39.36 \\
 
 & \multicolumn{1}{c|}{\multirow{-2}{*}{IMV}} & UVaT\pub{TIP24}~\cite{10666988}       &0.00      &0.00      &68.37       & 19.89      &40.87       &38.30       \\ \cline{2-3}
\multirow{-9}{*}{VGGSound} &
  \multicolumn{2}{c|}{CSCF (Ours)} &
  \textbf{8.74} &
  \textbf{8.65} &
  \textbf{69.31} &
  \textbf{60.97} &
  \textbf{64.58} &
  \textbf{45.10} \\ 
  \hline
  \hline
\end{tabular}
}
\end{table*}

\section{Experiments}
We conduct an in-depth evaluation to the proposed method, including real-world practicality, generalization ability, multimodal fusion ability, ablation study, uncertainty analysis, and hyperparameter analysis. 
\subsection{Experimental Setup}
\label{}
\textbf{Dataset.}
We conduct experiments on three multimodal audiovisual datasets: UCF, ActivityNet, and VGGSound~\cite{kurzendorfer2024audio}. For each dataset, we evenly divide the classes into two parts: the first part serves as seen classes for the audio modality (denoted $A_s$), while the second part serves as seen classes for the video modality (denoted $B_s$), maintaining the relationship where Classes($A_s$)=Classes($B_u$) and Classes($A_u$)=Classes($B_s$). During the training phase, the model exclusively utilizes data from $A_s$ of the audio modality and $B_s$ of the video modality for training.

\noindent\textbf{Comparison.}
Seven state-of-the-art (SOTA) methods are selected for comparison with our proposed CSCF model. These methods include: Unimodal Zero-Shot Methods~\cite{chen2022zero,chen2023deconstructed,li2024improving} , Multimodal Classification Methods~\cite{han2020trusted,han2022multimodal} and Incomplete Muti-View Methods~\cite{liu2023incomplete,10666988}. For Unimodal Zero-Shot Methods, since multimodal capability is unsupported, we use a dual-model architecture: independent unimodal models are trained for modality A and modality B. During testing, the appropriate model is selected based on the input modality, and the result with higher confidence from the two models is chosen for multimodal inputs.

\noindent\textbf{Implement Details.}
 Implementation details are provided in Appendix D. The main codes and logs are provided in Supplementary Material.

\begin{table}[t]
\caption{Gradual improvement in the model's fusion ability as the DMSS and CSMF modules are progressively added.}
\label{tab:ablation DMSS CSMF}
\centering
\begin{tabular}{l|l|lll}
\hline
\hline
Dataset & Model & \( A_s + B_u \) & \( A_u + B_s \) & \( A_{\text{all}} + B_{\text{all}} \) \\ \hline
        & B+O   & 45.70       & 82.39       & 65.15           \\
        & B+O+D & 47.01\textcolor{ForestGreen}{(↑)}    & 84.10\textcolor{ForestGreen}{(↑)}     & 66.76\textcolor{ForestGreen}{(↑)}        \\
\multirow{-3}{*}{ActivityNet} &
  B+O+D+C &
  \textbf{48.61\textcolor{ForestGreen}{(↑)}} &
 \textbf{84.33\textcolor{ForestGreen}{(↑)}} &
  \textbf{67.55\textcolor{ForestGreen}{(↑)}} \\ \hline
        & B+O   & 83.83       & 90.07       & 86.94           \\
        & B+O+D & 85.93\textcolor{ForestGreen}{(↑)}    & 92.88\textcolor{ForestGreen}{(↑)}    & 89.39\textcolor{ForestGreen}{(↑)}        \\
\multirow{-3}{*}{UCF} &
  B+O+D+C &
\textbf{86.23\textcolor{ForestGreen}{(↑)}} &
\textbf{93.08\textcolor{ForestGreen}{(↑)}} &
\textbf{89.64\textcolor{ForestGreen}{(↑)}} \\ \hline
        & B+O   & 66.24       & 58.77       & 60.36           \\
        & B+O+D & 68.87\textcolor{ForestGreen}{(↑)}    & 60.45\textcolor{ForestGreen}{(↑)}    & 63.96\textcolor{ForestGreen}{(↑)}        \\
\multirow{-3}{*}{VGGSound} &
  B+O+D+C &
\textbf{69.31\textcolor{ForestGreen}{(↑)}} &
\textbf{60.97\textcolor{ForestGreen}{(↑)}} &
\textbf{64.58\textcolor{ForestGreen}{(↑)}} \\ 
  \hline
  \hline
\end{tabular}
\end{table}

\begin{figure}[t]
    \centering
    \includegraphics[width=0.5\linewidth]{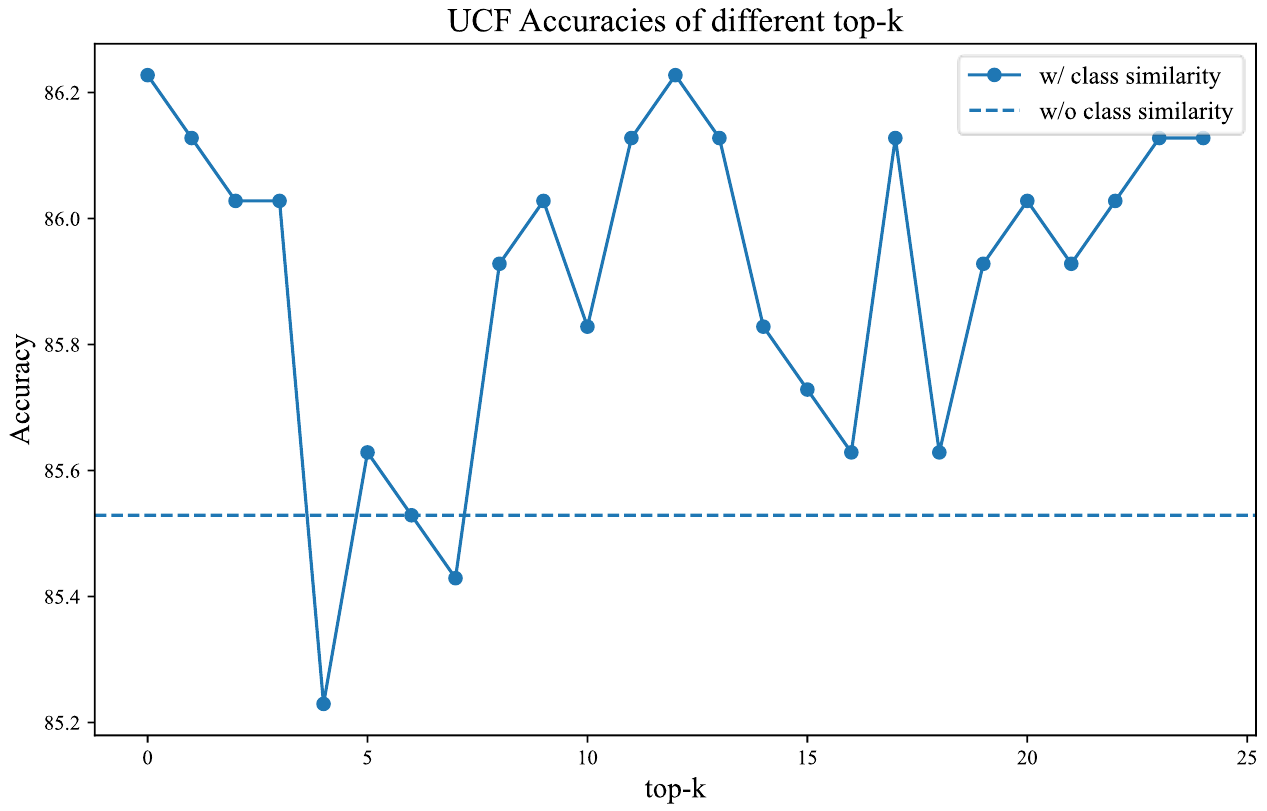}
    \caption{Analysis of the hyperparameter $top$-$k$ on the UCF dataset. "w/o class similarity" means no similarity fusion is used, while "w/ class similarity" indicates that similarity fusion is applied. We gradually increase the $top$-$k$ to observe its impact on the model's fusion capability.}
    \label{fig:hyperparameter analysis}
\end{figure}

\subsection{Comparison with SOTA}
\label{}
We conduct a comprehensive comparison between our model and the state-of-the-art (SOTA) unimodal zero-shot methods, multimodal classification methods, and incomplete multi-view methods as follows.

\noindent\textbf{Comprehensive Evaluation in Real-World Scenarios.}

As shown in~\Cref{tab:comparison method} (Real-world Scenarios). It can be observed that our method outperforms the best SOTA method by $10.23$ points on ActivityNet dataset, demonstrating that our model exhibits stable predictive performance when dealing with diverse modality combinations (unimodal/multimodal) and category distributions (seen classes/unseen classes) in the input data.

\noindent\textbf{Generalization Ability Evaluation.}

As shown in~\Cref{tab:comparison method} (Generalization Ability). On the one hand, multimodal classification and incomplete view methods typically use fully connected layers as classification heads, which causes the model to be biased toward seen classes, resulting in a lack of generalization ability. On the other hand, our approach introduces large model-enhanced semantic representations as a bridge for cross-category knowledge transfer, establishing effective semantic associations between seen and unseen classes via distance metrics, significantly improving our model's zero-shot capability.

\noindent\textbf{Multimodal Fusion Ability Evaluation.}

As shown in~\Cref{tab:comparison method} (Multimodal Fusion Ability). We conclude that: 1) Unseen class modality noise can disrupt model's decision-making, despite containing useful information; 2) The unimodal zero-shot methods that adopt an independent dual-model architecture process multimodal data by using a confidence selection strategy. However, unseen class modalities may produce overconfident predictions (see Appendix E for details), making it difficult to accurately identify the dominant modality; 3) Multimodal classification and incomplete view methods, due to the lack of cross-modal alignment supervision during training, struggle to establish effective inter-modal relationships; 4) Our method reliably identifies modality domains and dominant modalities, suppressing noise while leveraging similarity measures to extract useful unseen-class information for improved performance.

\subsection{Ablation Study}
\label{}

\begin{figure*}[t]
    \centering
    \includegraphics[width=\linewidth]{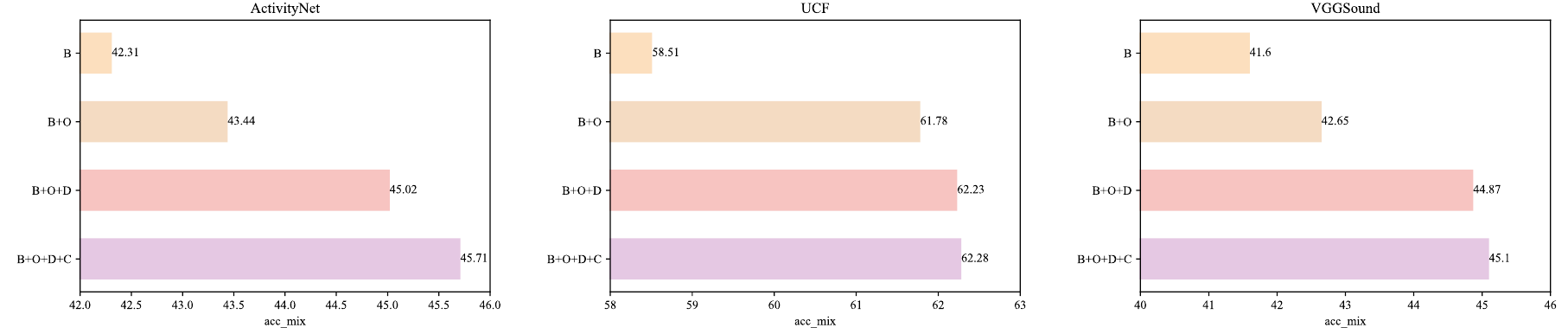}
    \caption{The impact of OSRS(O), DMSS(D), and CSMF(C) on the comprehensive performance of the model. We remove these three modules as our Baseline (B). In the ablation study, we add O, D, and C step by step to demonstrate their influence on the model's overall performance.}
    \label{fig:ablation experiments}
\end{figure*}

Since our three modules each affect different functional aspects of the model, the experiment first quantitatively evaluates the contribution of each module to the overall performance in real-world scenarios, followed by a deeper analysis of the effectiveness of each module in specific functional dimensions. As shown in ~\Cref{fig:ablation experiments}, we observe that as the OSRS, DMSS, and CSMF modules are progressively integrated into the baseline model, the overall performance in real-world scenarios gradually improves, which strongly demonstrates the effectiveness of each module in enhancing the overall performance in real-world scenarios.

\begin{figure}[!t]
    \centering
    \includegraphics[width=0.5\linewidth]{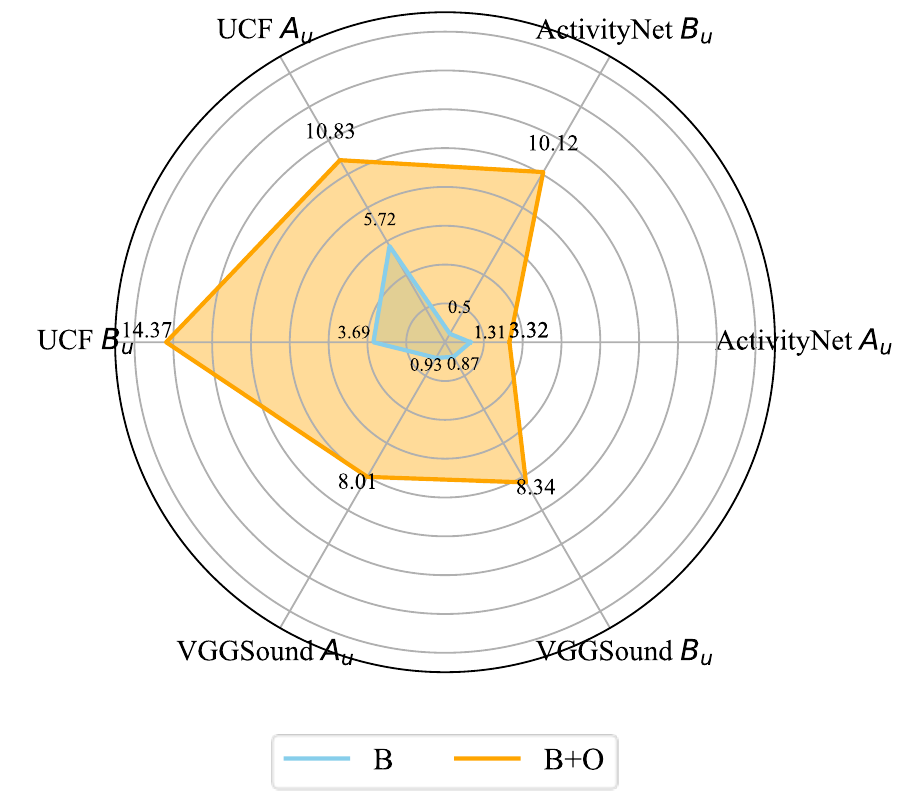}
    \caption{Comparison of the model's ability to recognize unseen classes before and after introducing the OSRS module."B+O" means adding the OSRS module to the baseline.}
    \label{fig:ablation_B+O}
\end{figure}

\noindent\textbf{The effectiveness of the OSRS module on model generalization.}

As shown in ~\Cref{fig:ablation_B+O}, the OSRS module, compared to the traditional fully connected + Softmax classification method, significantly enhances the model’s ability to recognize unseen categories through the semantic space alignment mechanism. Additionally, by incorporating semantic enhancement strategies, this module makes the category prototypes more representative within the shared semantic space, thereby further improving the model's generalization capability.

\noindent\textbf{The effectiveness of the DMSS module on model fusion capability.}

As shown in ~\Cref{tab:ablation DMSS CSMF}, compared to the simple modality prediction averaging fusion strategy, the DMSS module dynamically selects the dominant modality's prediction result through adaptive inconsistency estimation. This mechanism effectively suppresses the noise interference from unseen class data and enhances the model's robustness under MMHCL scenario.

\noindent\textbf{The effectiveness of the CSMF module on model fusion capability.}
As shown in ~\Cref{tab:ablation DMSS CSMF}, directly selecting the dominant modality's prediction result (without fusion) leads to the loss of some useful information. However, the CSMF module, through cross-modal similarity-weighted fusion, effectively utilizes the beneficial information from the auxiliary modality, thereby further enhancing the accuracy of multimodal decision-making.

\subsection{Experimental Analysis}
\label{experiment:uncertainty analysis}
\begin{figure}[t]
    \centering
    \includegraphics[width=\linewidth]{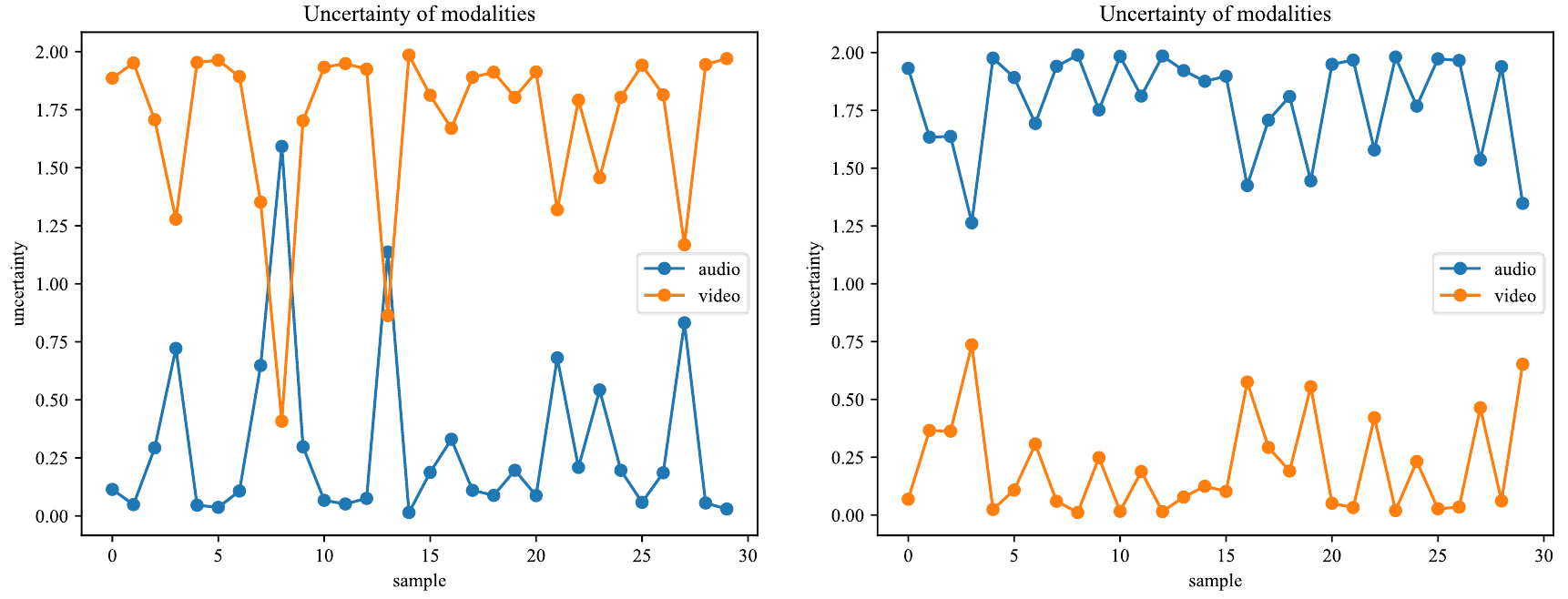}
    \caption{Comparison of prediction uncertainty across modalities. The horizontal axis represents 30 randomly selected samples, and the vertical axis represents the prediction uncertainty of each modality for the corresponding sample. The left subfigure presents the test results on the seen classes of the audio modality, while the right subfigure shows the test results on the seen classes of the video modality.}
    \label{fig:uncertainty analysis}
\end{figure}

\textbf{Comparison of Modality Prediction Uncertainty.}
By combining the intra-modal prediction inconsistency $Inc^*$
mentioned in~\Cref{sec:DMSS} with the inter-modal prediction discrepancy $dif^*$, we obtain the final modal prediction uncertainty $u^*$. We visualize the prediction uncertainty of each modality, as shown in~\Cref{fig:uncertainty analysis}. It can be observed that, across 30 randomly selected samples, each modality exhibits low prediction uncertainty for its own seen classes, demonstrating high confidence in predictions. This suggests that these modalities can serve as the dominant ones in subsequent multimodal decision fusion.

\noindent\textbf{Hyperparameter Analysis.}
As shown in ~\Cref{fig:hyperparameter analysis}, as $top$-$k$ becomes smaller, the similarity weighting becomes more focused on the most relevant few categories. While increasing $top$-$k$ introduces more information from the unseen class modalities, it also brings in more noise.

\section{Conclusion}
In this paper, we propose an effective approach that aligns different modalities into a unified semantic space while integrating ensemble learning and class similarity, enabling a decision-making process driven by the dominant modality with auxiliary support from other modalities. From the experimental results, we draw the following key conclusions: 1) The MMHCL scenario is both practically significant and challenging. Existing generalized zero-shot learning methods, multimodal classification approaches, and incomplete-view methods experience significant performance degradation in this scenario; 2) 
Our model achieves strong generalization while effectively leveraging information from different modalities, performing well across various scenarios with a single training process. 
In the future, we will explore MMHCL methods based on federated learning, leveraging multi-center data while preserving privacy to further enhance model performance.

\noindent\textbf{Potential Impact.} The paper will help attract researchers' attention to MMHCL and promote the development of multimodal learning toward a more practical direction. The proposed CSCF, by effectively integrating data from different projects and modalities, demonstrates outstanding generalization and multimodal fusion capabilities with just one training session. This property enables companies to maximize the use of data from various past projects, whether historical or cross-domain, providing rich training information for the model. As a result, the performance and robustness of the model are significantly enhanced, ultimately leading to the training of a more powerful model.

\newpage

\bibliographystyle{unsrt}  
\bibliography{ref}
\newpage

\appendix
\section{Semantic Enhancement}
\label{appendix:SA}

\begin{table*}[!t]\small
    \centering
    \caption{Examples of class description generated by SA. The \textcolor{blue}{blue} and \textcolor{purple}{purple} sentences are audio and video feature.}
    \begin{tabular}{c|c}
     \hline
      Original Class Name   &  Class Description\\ \hline \hline
      Swimming
 & \textcolor{blue}{The rhythmic splash of strokes and the occasional gasp for air.} \textcolor{purple}{A swimmer cutting through the water.}
     \\
     Cat purring & \textcolor{blue}{The soothing, vibrating hum of a content cat.}\textcolor{purple}{A relaxed feline curling up in a cozy spot.}
 \\
      
     Cleaningshoes&\textcolor{blue}{ The squeak of a brush against leather. }\textcolor{purple}{  Shoes being polished to a glossy shine.
}\\
     Knitting&\textcolor{blue}{The click-clack of knitting needles and the soft pull of yarn.}\textcolor{purple}{ A colorful scarf or sweater taking shape on someone’s lap.}\\
     \hline
    \end{tabular}
    
    \label{tab:enhanced semantics}
\end{table*}

We have designed a prompt-based strategy to guide the large language model (LLM) in transforming category names into descriptive statements that capture their attributes, thereby generating richer and more informative semantic features. The generated statements are then used as input to extract semantic features. Formally, the extraction of enhanced semantic features for category names is as follows:
\begin{equation}
\begin{split}
    \label{eq:semantic enhancement}
    c_i = CLIP(LLM(Y_i))
\end{split}
\end{equation}

Using a large language model (LLM), we transform the original category names into descriptive sentences, as shown in Table \ref{tab:enhanced semantics}. It can be observed that, compared to the raw category names, the sentence descriptions contain richer visual and auditory information, which helps enhance the transfer from seen to unseen categories.

\begin{figure}[!t]
    \centering
    \includegraphics[width=0.5\linewidth]{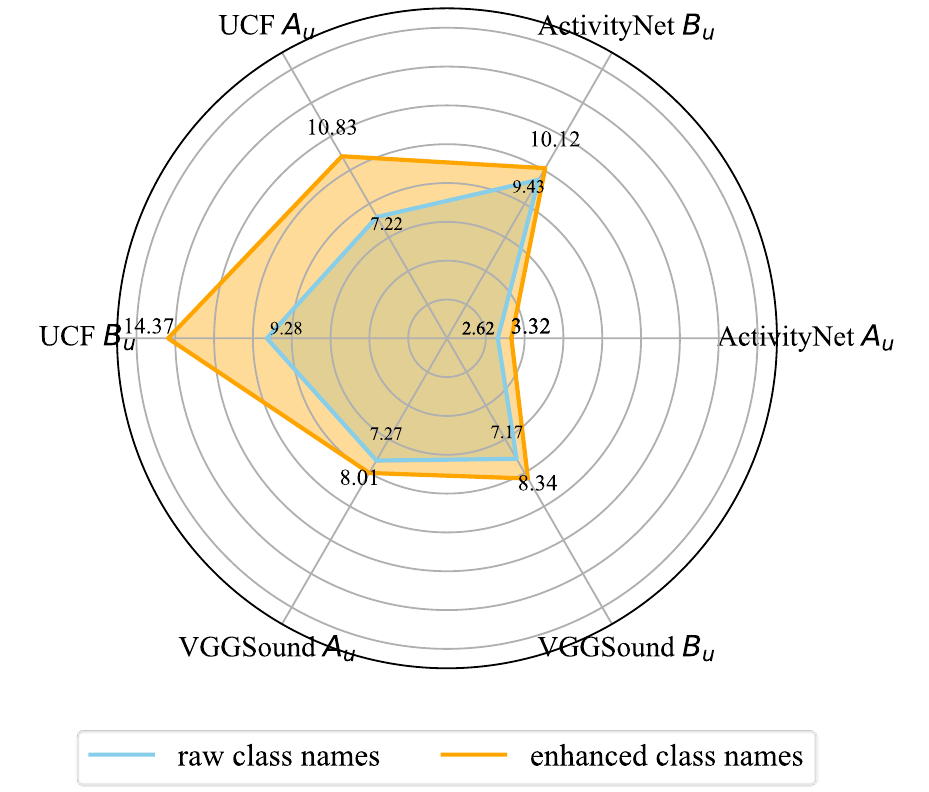}
    \caption{Comparison of the model(B+O)’s ability to recognize unseen classes before and after incorporating semantic enhancement.}
    \label{fig:ablation SA}
\end{figure}

We conduct an ablation study on the semantic enhancement module based on the B+O model, as shown in \Cref{fig:ablation SA}. It can be observed that incorporating semantic enhancement significantly improves the model’s ability to recognize unseen classes compared to the version without semantic enhancement.

\section{The Inconsistency in Intra-Modal Predictions}
\label{appendix:inconsistency of predictions}
\begin{figure}[htb]
    \centering
    \includegraphics[width=\linewidth]{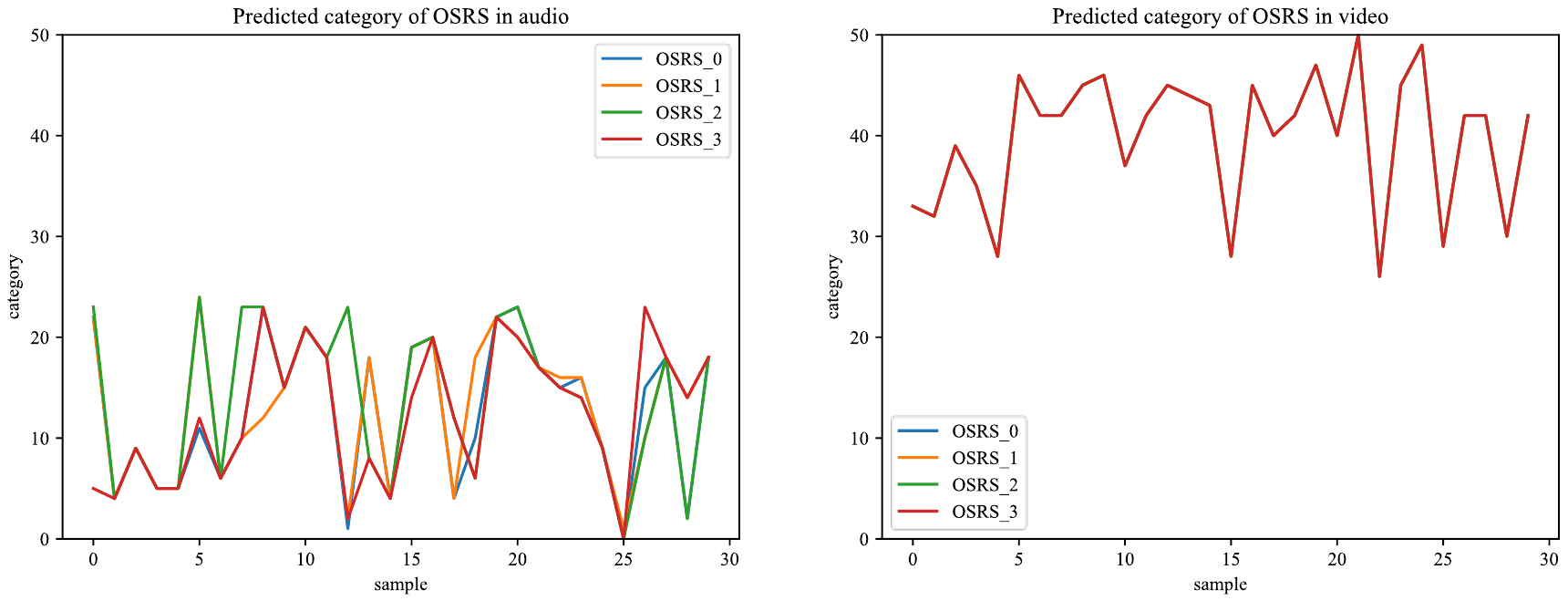}
    \caption{The predictions of each OSRS module within individual modalities are illustrated. The x-axis represents the 30 randomly selected samples, while the y-axis indicates the predicted categories.}
    \label{fig:inconsistency of predictions}
\end{figure}

We adopt an ensemble learning strategy by training four OSRS modules with different architectures within each modality and randomly select 30 samples for testing. As shown in Figure \ref{fig:inconsistency of predictions}, each polyline represents the predictions from a sub-network for the 30 samples. The right plot displays predictions for seen classes in the video modality, where the high overlap of polylines indicates strong prediction consistency among the four OSRS modules. In contrast, the left plot shows predictions for unseen classes in the audio modality, with less polyline overlap, revealing lower consistency across the modules.

\section{Evaluation Methods}
We evaluate three key aspects: real-world applicability, generalization ability (for Challenge 1), and multimodal fusion ability (for Challenges 2-3).
\begin{figure}[t]
    \centering
    \includegraphics[width=0.5\linewidth]{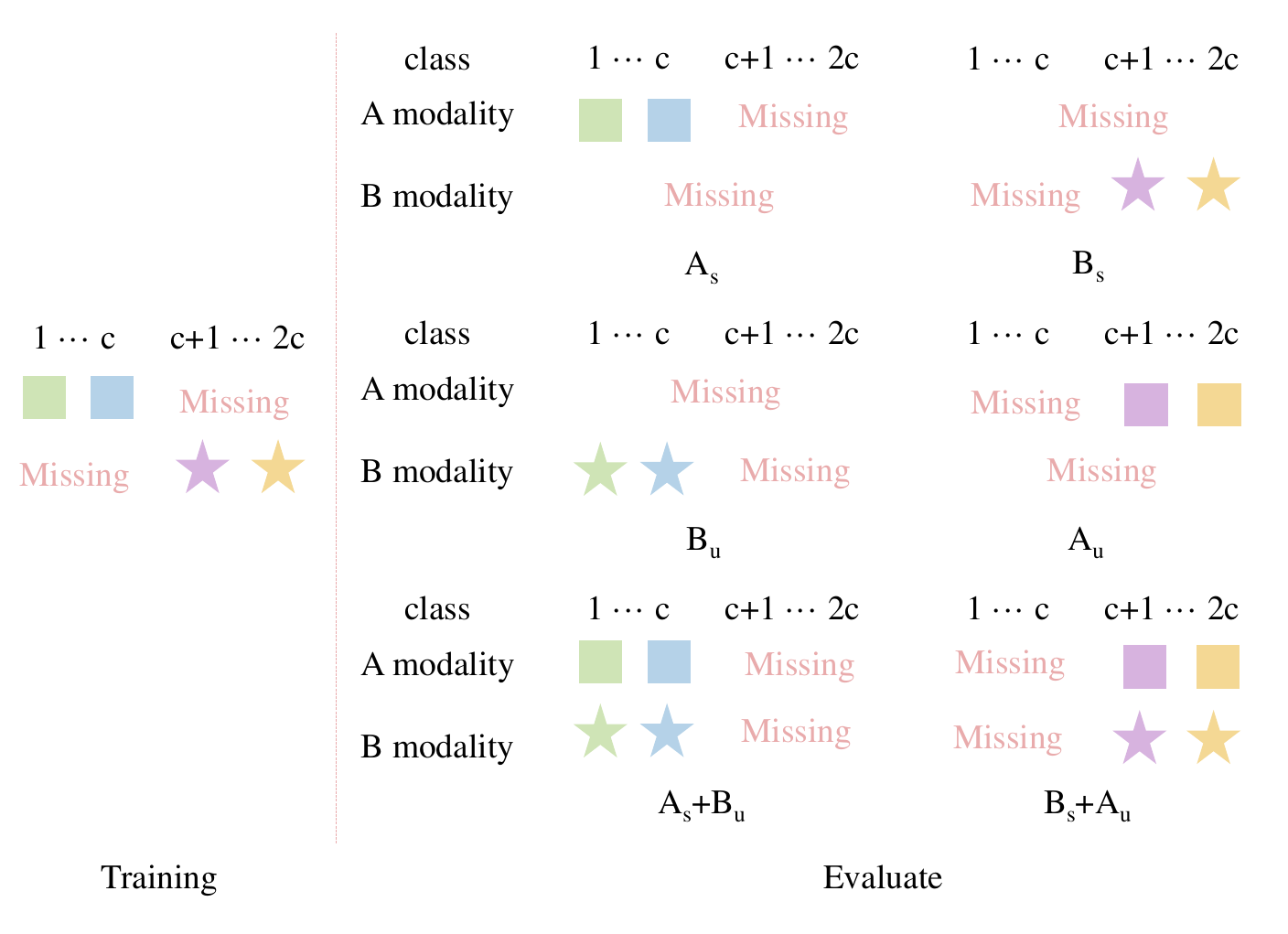}
    \caption{Visualization of the existence states of multimodal data during training and evaluate.}
    \label{fig:evaluate}
\end{figure}

\noindent\textbf{Real-world Scenarios Evaluation}
\begin{itemize}
    \item Unimodal seen class testing ($A_s$, $B_s$)
    \item Unimodal unseen class testing ($A_u$, $B_u$)
    \item Cross-modal mixed testing ($A_s$ + $B_u$, $B_s$ + $A_u$)
\end{itemize}

\noindent\textbf{Generalization Ability Evaluation}
\begin{itemize}
    \item Modality A unseen class testing ($A_u$)
    \item Modality B unseen class testing ($B_u$)
\end{itemize}

\noindent\textbf{Multimodal Fusion Ability Evaluation}

\begin{itemize}
    \item Modality A seen class + Modality B unseen class ($A_s$ + $B_u$)
    \item Modality B seen class + Modality A unseen class ($B_s$ + $A_u$)
\end{itemize}
The diagrams of $A_s$, $B_s$, $A_u$, $B_u$, $A_s$ + $B_u$, $B_s$ + $A_u$ are shown in~\Cref{fig:evaluate}.

\section{Implementation Details}
\label{appendix:implementation details}
For the audio and visual features of ActivityNet, UCF, and VGGSound ($d^{a}=1024$ and $d^{v}=512$) ,we use features extracted by CLIPCLAP~\cite{kurzendorfer2024audio}.Missing modality data is padded with zero vectors (1024-D for audio and 512-D for video). The semantic embeddings of the classes are obtained using the CLIP ViT-B/32 text encoder ($d^{s}=512$). We train our model using the Adam optimizer with a learning rate of $5 \times 10^{-4}$ and weight decay of $10^{-4}$. The training epoch and batch size are set to 50 and 256, respectively. Our model is implemented in PyTorch. All experiments are conducted under the same hardware environment, including an NVIDIA GeForce RTX 4090 GPU,an Intel Core i9-13900K CPU, and 128 GB of RAM.

\section{Confidence Analysis}
\label{appandix:other analysis}

\begin{figure}[!t]
    \centering
    \includegraphics[width=0.5\linewidth]{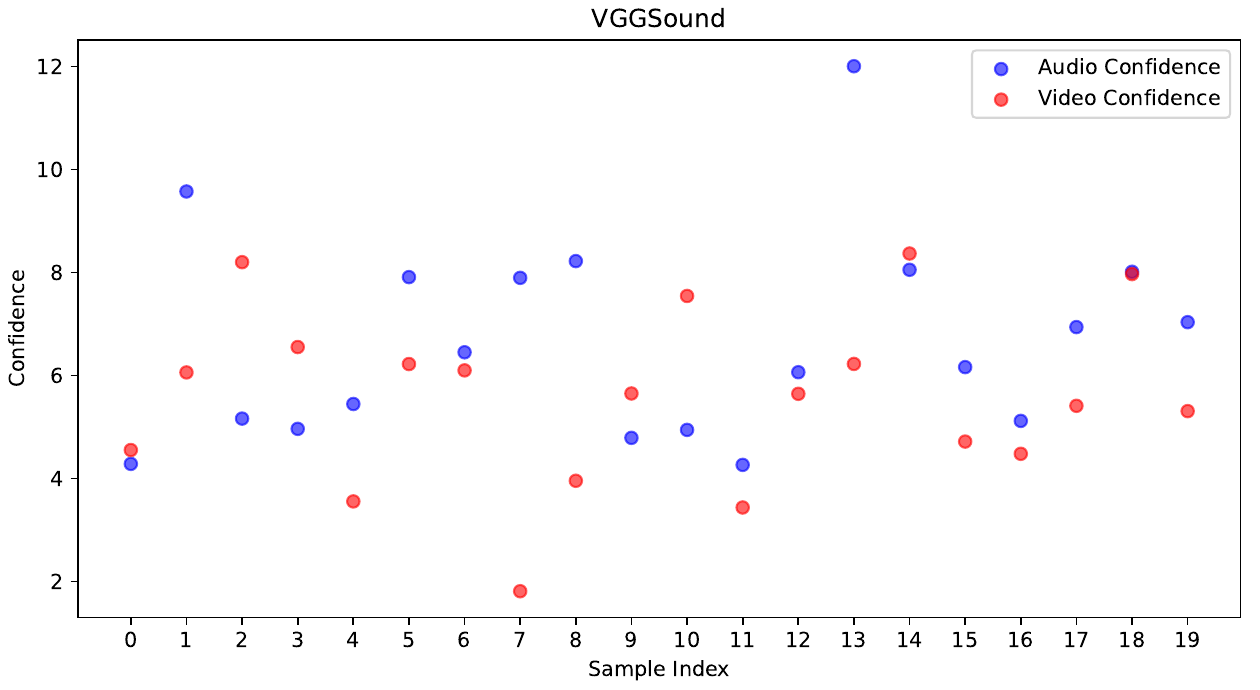}
    \caption{Confidence levels of two zero-shot models trained separately on audio and video modalities during full-modality testing. Despite the audio modality not having encountered these classes, certain samples still exhibit high audio confidence. Blue dots indicate audio confidence, while red dots indicate video confidence.}
    \label{fig:confidence analysis}
\end{figure}

As shown in ~\Cref{fig:confidence analysis}, for two zero-shot models trained separately on audio and video modalities,  we randomly sampled twenty instances from classes that were only seen in the video modality during training. The results reveal that although these classes were unseen in the audio modality, the model still produces over confident predictions for certain samples.

\end{document}